\documentclass[10pt,twocolumn,letterpaper]{article}

\usepackage{cvpr}
\usepackage{times}
\usepackage{epsfig}
\usepackage{graphicx}
\usepackage{amsmath}
\usepackage{amssymb}
\usepackage{bbm}
\usepackage{enumitem, url}
\usepackage{comment}
\usepackage{tabulary, multirow}

\usepackage{color}

\newcolumntype{x}[1]{>{\centering\arraybackslash}p{#1pt}}
\newlength\savewidth\newcommand\shline{\noalign{\global\savewidth\arrayrulewidth
  \global\arrayrulewidth 1pt}\hline\noalign{\global\arrayrulewidth\savewidth}}
\newcommand{\tablestyle}[2]{\setlength{\tabcolsep}{#1}\renewcommand{\arraystretch}{#2}\centering\footnotesize}

\newcommand{\bd}[1]{\textbf{#1}}
\newcommand{\app}{\raise.17ex\hbox{$\scriptstyle\sim$}}

\def\x{\times}

\usepackage[pagebackref=true,breaklinks=true,letterpaper=true,colorlinks,bookmarks=false]{hyperref}

\cvprfinalcopy 

\def\cvprPaperID{1352} 

\ifcvprfinal\fi
\begin{document}

\title{YOLOv3: An Incremental Improvement}

\author{Joseph Redmon \quad Ali Farhadi\\
{\normalsize University of Washington}
}
\maketitle

\begin{abstract}

We present some updates to YOLO! We made a bunch of little design changes to make it better. We also trained this new network that's pretty swell. It's a little bigger than last time but more accurate. It's still fast though, don't worry. At $320\times320$ YOLOv3 runs in 22 ms at 28.2 mAP, as accurate as SSD but three times faster. When we look at the old .5 IOU mAP detection metric YOLOv3 is quite good. It achieves $57.9$ AP$_{50}$ in 51 ms on a Titan X, compared to $57.5$ AP$_{50}$ in 198 ms by RetinaNet, similar performance but 3.8$\x$ faster. As always, all the code is online at \url{https://pjreddie.com/yolo/}.

\end{abstract}

\section{Introduction}

Sometimes you just kinda phone it in for a year, you know? I didn't do a whole lot of research this year. Spent a lot of time on Twitter. Played around with GANs a little. I had a little momentum left over from last year \cite{newton1687} \cite{analogy}; I managed to make some improvements to YOLO. But, honestly, nothing like super interesting, just a bunch of small changes that make it better. I also helped out with other people's research a little.

Actually, that's what brings us here today. We have a camera-ready deadline \cite{gordon2017iqa} and we need to cite some of the random updates I made to YOLO but we don't have a source. So get ready for a TECH REPORT!

The great thing about tech reports is that they don't need intros, y'all know why we're here. So the end of this introduction will signpost for the rest of the paper. First we'll tell you what the deal is with YOLOv3. Then we'll tell you how we do. We'll also tell you about some things we tried that didn't work. Finally we'll contemplate what this all means.

\section{The Deal}

So here's the deal with YOLOv3: We mostly took good ideas from other people. We also trained a new classifier network that's better than the other ones. We'll just take you through the whole system from scratch so you can understand it all.

\begin{figure}[t]
\hspace{-6mm}
\includegraphics[width=1.06\linewidth]{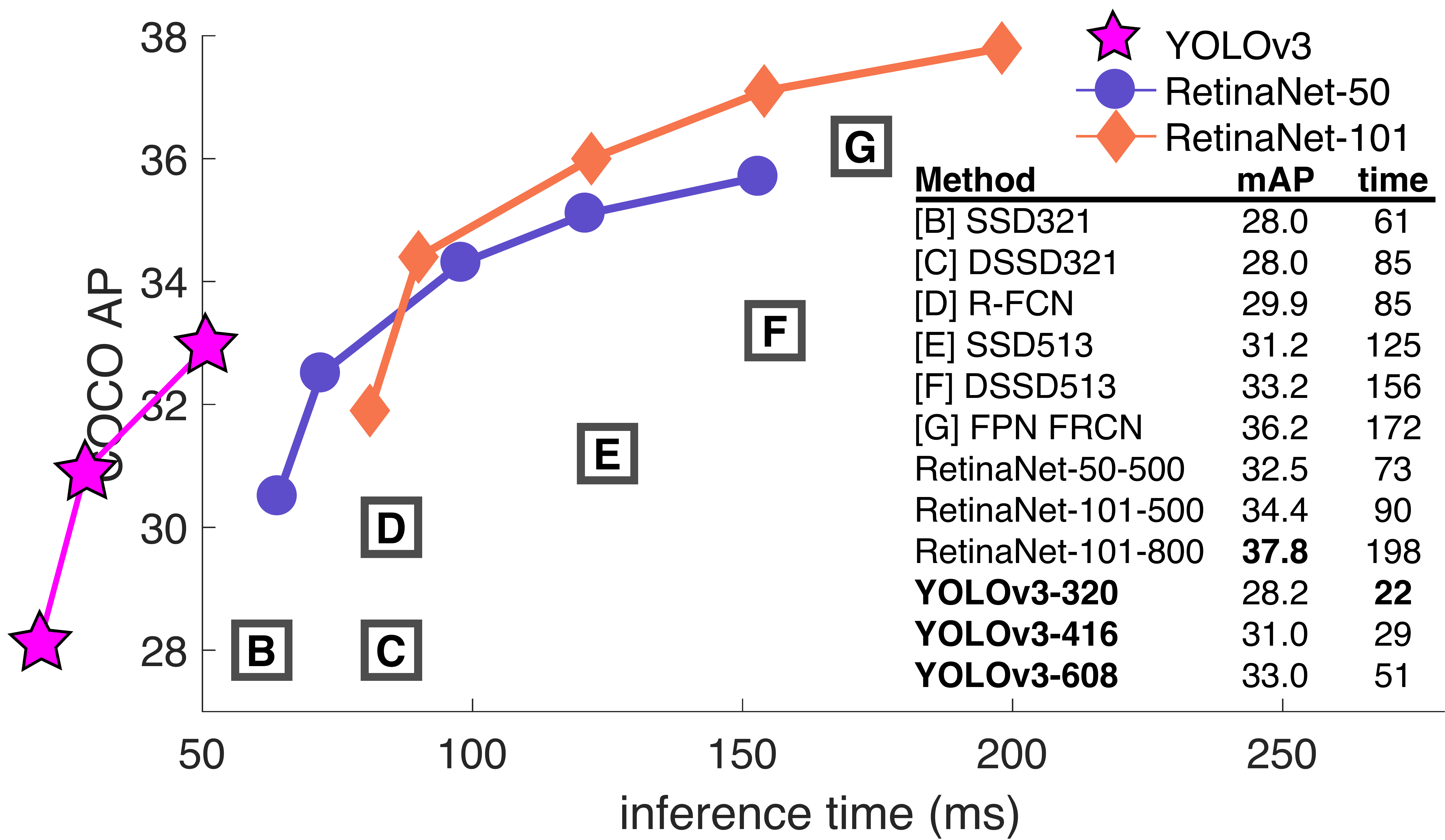}
\caption{We adapt this figure from the Focal Loss paper \cite{focal}. YOLOv3 runs significantly faster than other detection methods with comparable performance. Times from either an M40 or Titan X, they are basically the same GPU.}
\label{fig:teaser}
\vspace{-4mm}
\end{figure}

\subsection{Bounding Box Prediction}

Following YOLO9000 our system predicts bounding boxes using dimension clusters as anchor boxes \cite{redmon2017yolo9000}. The network predicts 4 coordinates for each bounding box, $t_x$, $t_y$, $t_w$, $t_h$. If the cell is offset from the top left corner of the image by $(c_x, c_y)$ and the bounding box prior has width and height $p_w$, $p_h$, then the predictions correspond to:

\begin{align*}
b_x &= \sigma(t_x) + c_x \\
b_y &= \sigma(t_y)  + c_y\\
b_w &= p_w e^{t_w}\\
b_h &= p_h e^{t_h}\\
\end{align*}

During training we use sum of squared error loss. If the ground truth for some coordinate prediction is $\hat{t}_{\mbox{*}}$ our gradient is the ground truth value (computed from the ground truth box) minus our prediction: $\hat{t}_{\mbox{*}} - t_{\mbox{*}}$. This ground truth value can be easily computed by inverting the equations above.

\begin{figure}[]
      \centering
        \includegraphics[width=\linewidth]{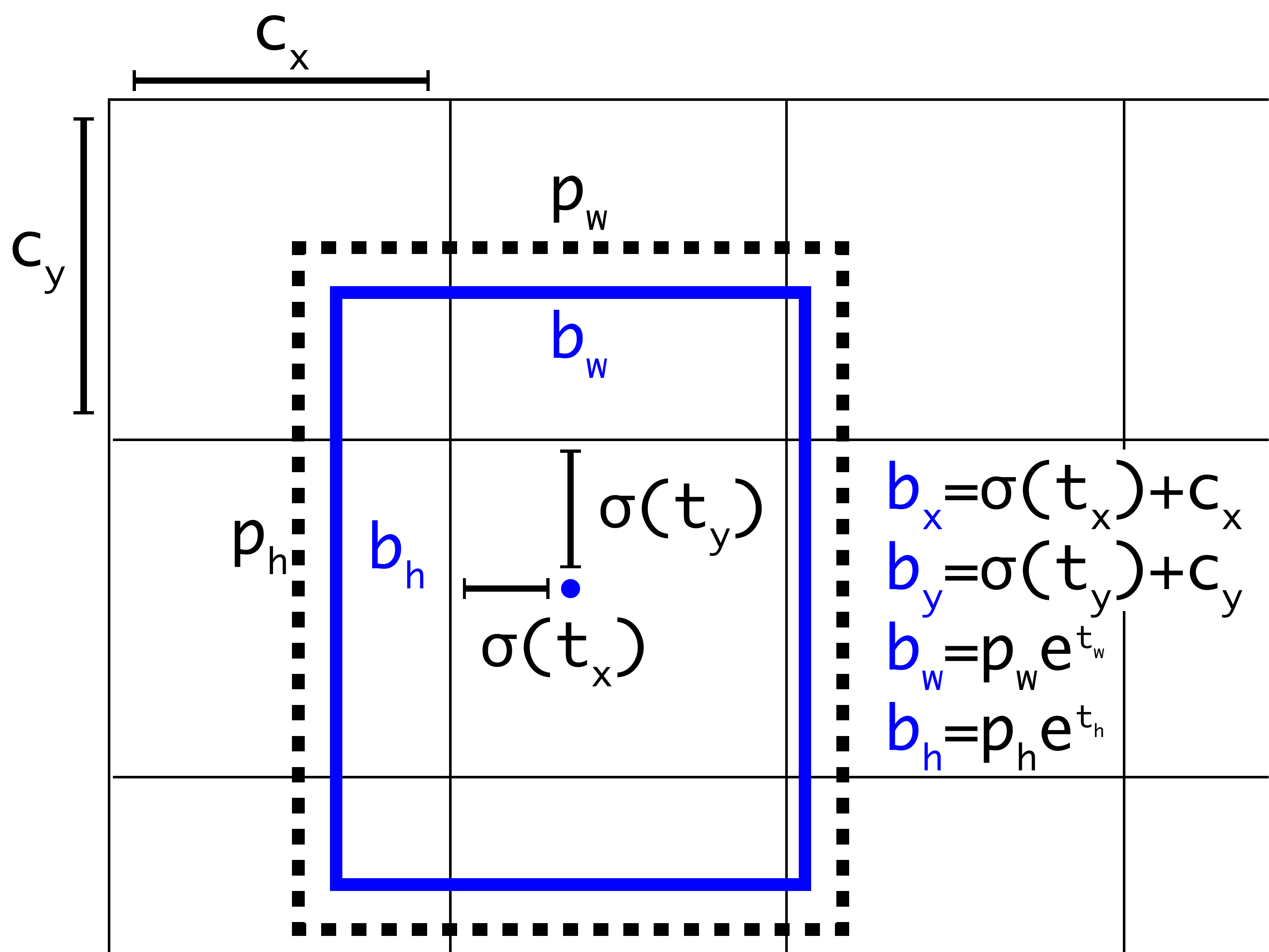}
      \caption{\small \textbf{Bounding boxes with dimension priors and location prediction.} We predict the width and height of the box as offsets from cluster centroids. We predict the center coordinates of the box relative to the location of filter application using a sigmoid function. This figure blatantly self-plagiarized from \cite{redmon2017yolo9000}.}
      \label{box}
   \end{figure}

YOLOv3 predicts an objectness score for each bounding box using logistic regression. This should be 1 if the bounding box prior overlaps a ground truth object by more than any other bounding box prior. If the bounding box prior is not the best but does overlap a ground truth object by more than some threshold we ignore the prediction, following \cite{ren2015faster}. We use the threshold of $.5$. Unlike \cite{ren2015faster} our system only assigns one bounding box prior for each ground truth object. If a bounding box prior is not assigned to a ground truth object it incurs no loss for coordinate or class predictions, only objectness.

\subsection{Class Prediction}

Each box predicts the classes the bounding box may contain using multilabel classification. We do not use a softmax as we have found it is unnecessary for good performance, instead we simply use independent logistic classifiers. During training we use binary cross-entropy loss for the class predictions.

This formulation helps when we move to more complex domains like the Open Images Dataset \cite{openimages}. In this dataset there are many overlapping labels (i.e. Woman and Person). Using a softmax imposes the assumption that each box has exactly one class which is often not the case. A multilabel approach better models the data.

\subsection{Predictions Across Scales}

YOLOv3 predicts boxes at 3 different scales. Our system extracts features from those scales using a similar concept to feature pyramid networks \cite{lin2017feature}. From our base feature extractor we add several convolutional layers. The last of these predicts a 3-d tensor encoding bounding box, objectness, and class predictions. In our experiments with COCO \cite{lin2014microsoft} we predict 3 boxes at each scale so the tensor is $N\times N\times [3*(4+1+80)]$ for the 4 bounding box offsets, 1 objectness prediction, and 80 class predictions.

Next we take the feature map from 2 layers previous and upsample it by $2\times$. We also take a feature map from earlier in the network and merge it with our upsampled features using concatenation. This method allows us to get more meaningful semantic information from the upsampled features and finer-grained information from the earlier feature map. We then add a few more convolutional layers to process this combined feature map, and eventually predict a similar tensor, although now twice the size.

We perform the same design one more time to predict boxes for the final scale. Thus our predictions for the 3rd scale benefit from all the prior computation as well as fine-grained features from early on in the network.

We still use k-means clustering to determine our bounding box priors. We just sort of chose 9 clusters and 3 scales arbitrarily and then divide up the clusters evenly across scales. On the COCO dataset the 9 clusters were: $(10 \times 13), (16\times 30), (33 \times 23), (30 \times 61), (62 \times 45), (59 \times 119), (116 \times 90), (156 \times 198), (373 \times 326)$.

\subsection{Feature Extractor}

We use a new network for performing feature extraction. Our new network is a hybrid approach between the network used in YOLOv2, Darknet-19, and that newfangled residual network stuff. Our network uses successive $3 \times 3$ and $1 \x 1$ convolutional layers but now has some shortcut connections as well and is significantly larger. It has 53 convolutional layers so we call it.... wait for it..... Darknet-53!

\begin{table}[h] 
\begin{center}
\includegraphics[width=.8\linewidth]{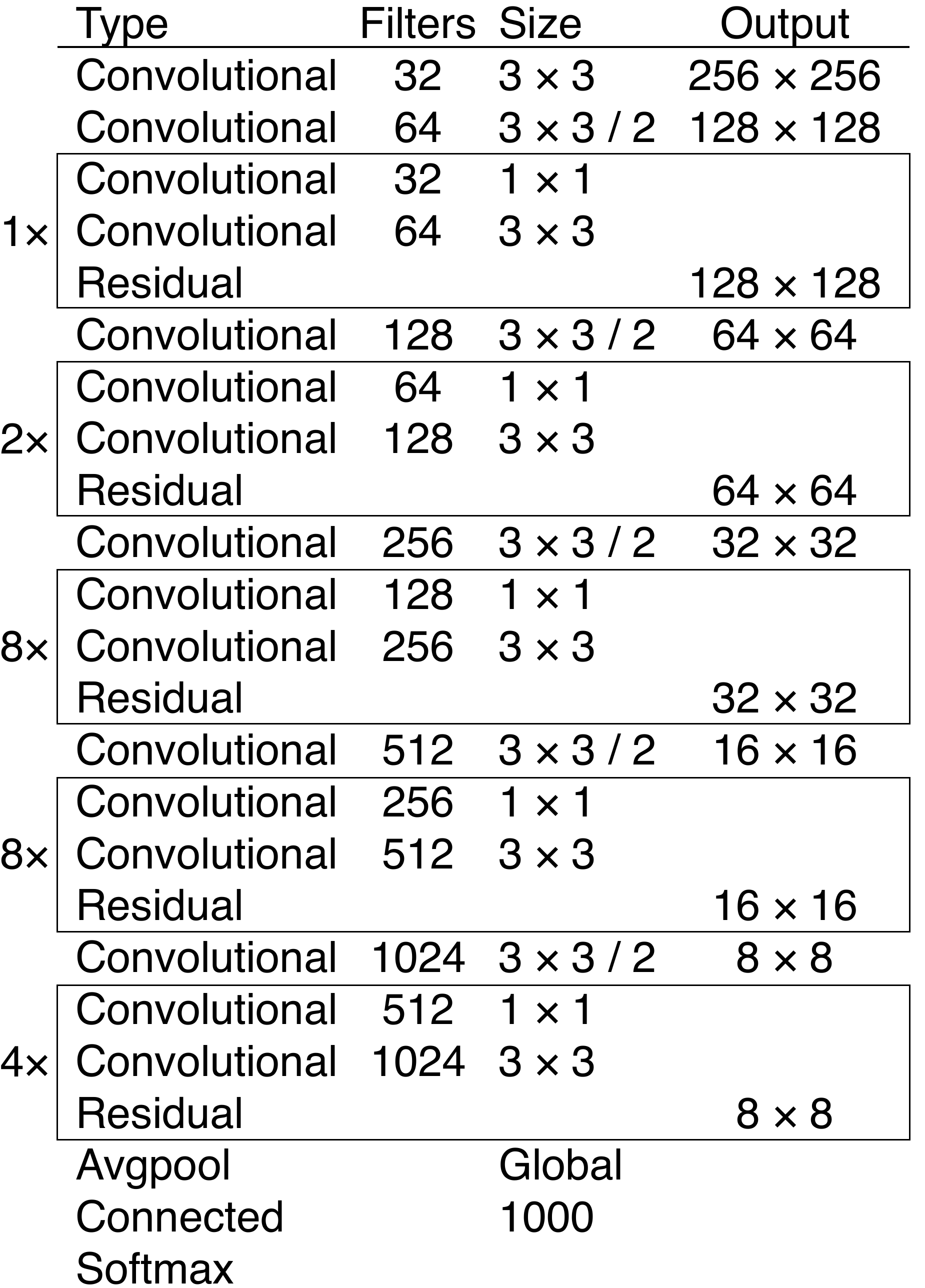}
\end{center}
\caption{\small \textbf{Darknet-53.}}
\label{net}
\end{table}

This new network is much more powerful than Darknet-19 but still more efficient than ResNet-101 or ResNet-152. Here are some ImageNet results:

\begin{table}[h]
\small
\begin{center}
\begin{tabular}{lrrrrr}
Backbone & Top-1 & Top-5 & Bn Ops & BFLOP/s & FPS\\
\hline
Darknet-19 \cite{redmon2017yolo9000}& 74.1 & 91.8 & 7.29 & 1246 & \bd{171}  \\
ResNet-101\cite{resnet}& 77.1 & 93.7 & 19.7 & 1039 & 53 \\
ResNet-152 \cite{resnet}& \bd{77.6} & \bd{93.8} & 29.4 & 1090 & 37 \\
Darknet-53 & 77.2 & \bd{93.8} & 18.7 & \bd{1457} & 78 \\
\end{tabular}
\end{center}
\caption{\small \textbf{Comparison of backbones.} Accuracy, billions of operations, billion floating point operations per second, and FPS for various networks.}
\label{imnet}
\vspace{-1mm}
\end{table}

Each network is trained with identical settings and tested at $256 \times 256$, single crop accuracy. Run times are measured on a Titan X at $256 \x 256$. Thus Darknet-53 performs on par with state-of-the-art classifiers but with fewer floating point operations and more speed. Darknet-53 is better than ResNet-101 and $1.5\x$ faster. Darknet-53 has similar performance to ResNet-152 and is $2\x$ faster.

Darknet-53 also achieves the highest measured floating point operations per second. This means the network structure better utilizes the GPU, making it more efficient to evaluate and thus faster. That's mostly because ResNets have just way too many layers and aren't very efficient.

\subsection{Training}

We still train on full images with no hard negative mining or any of that stuff. We use multi-scale training, lots of data augmentation, batch normalization, all the standard stuff. We use the Darknet neural network framework for training and testing \cite{darknet13}.

\section{How We Do}

\begin{table}[b]
\begin{minipage}{\textwidth}
\tablestyle{4pt}{1.05}
\begin{tabular}{l|c|x{22}x{22}x{22}|x{22}x{22}x{22}}
 & backbone
 & AP & AP$_{50}$ & AP$_{75}$
 & AP$_S$ & AP$_M$ &  AP$_L$\\ [.1em]
\shline
\emph{Two-stage methods} & & & & & & & \\
 ~Faster R-CNN+++ \cite{resnet} & ResNet-101-C4
  & 34.9 & 55.7 & 37.4 & 15.6 & 38.7 & 50.9\\
 ~Faster R-CNN w FPN \cite{lin2017feature} & ResNet-101-FPN
  & 36.2 & 59.1 & 39.0 & 18.2 & 39.0 & 48.2\\
 ~Faster R-CNN by G-RMI \cite{huang2017speed} & Inception-ResNet-v2 \cite{szegedy2017inception}
  & 34.7 & 55.5 & 36.7 & 13.5 & 38.1 & 52.0\\
 ~Faster R-CNN w TDM \cite{shrivastava2016beyond} & Inception-ResNet-v2-TDM
  & 36.8 & 57.7 & 39.2 & 16.2 & 39.8 & \bd{52.1}\\
\hline
\emph{One-stage methods} & & & & & & & \\
 ~YOLOv2 \cite{redmon2017yolo9000} & DarkNet-19 \cite{redmon2017yolo9000}
  & 21.6 & 44.0 & 19.2 & 5.0 & 22.4 & 35.5 \\
 ~SSD513 \cite{liu2016ssd,fu2017dssd} & ResNet-101-SSD
  & 31.2 & 50.4 & 33.3 & 10.2 & 34.5 & 49.8 \\
 ~DSSD513 \cite{fu2017dssd} & ResNet-101-DSSD
  & 33.2 & 53.3 & 35.2 & 13.0 & 35.4 & 51.1 \\
 ~RetinaNet \cite{focal} & ResNet-101-FPN
  & 39.1 & 59.1 & 42.3 & 21.8 & 42.7 & 50.2 \\
 ~RetinaNet \cite{focal} & ResNeXt-101-FPN
  & \bd{40.8} & \bd{61.1} & \bd{44.1} & \bd{24.1} & \bd{44.2} & 51.2 \\
  ~YOLOv3 $608 \times 608$ & Darknet-53
  & 33.0 & 57.9 & 34.4 & 18.3 & 35.4 & 41.9 \\
\end{tabular}
\vspace{1mm}
\caption{I'm seriously just stealing all these tables from \cite{focal} they take soooo long to make from scratch. Ok, YOLOv3 is doing alright. Keep in mind that RetinaNet has like $3.8\x$ longer to process an image. YOLOv3 is much better than SSD variants and comparable to state-of-the-art models on the AP$_{50}$ metric.}
\label{results}
\end{minipage}
\end{table}

YOLOv3 is pretty good! See table \ref{results}. In terms of COCOs weird average mean AP metric it is on par with the SSD variants but is $3\x$ faster. It is still quite a bit behind other models like RetinaNet in this metric though.

\enlargethispage{-15\baselineskip}

However, when we look at the ``old'' detection metric of mAP at IOU$=.5$ (or AP$_{50}$ in the chart) YOLOv3 is very strong. It is almost on par with RetinaNet and far above the SSD variants. This indicates that YOLOv3 is a very strong detector that excels at producing decent boxes for objects. However, performance drops significantly as the IOU threshold increases indicating YOLOv3 struggles to get the boxes perfectly aligned with the object.

In the past YOLO struggled with small objects. However, now we see a reversal in that trend. With the new multi-scale predictions we see YOLOv3 has relatively high AP$_S$ performance. However, it has comparatively worse performance on medium and larger size objects. More investigation is needed to get to the bottom of this.

When we plot accuracy vs speed on the AP$_{50}$ metric (see figure \ref{chart}) we see YOLOv3 has significant benefits over other detection systems. Namely, it's faster and better.

\begin{figure*}[]
\hspace{-11mm}
\vspace{-8mm}

\includegraphics[width=1.16\linewidth]{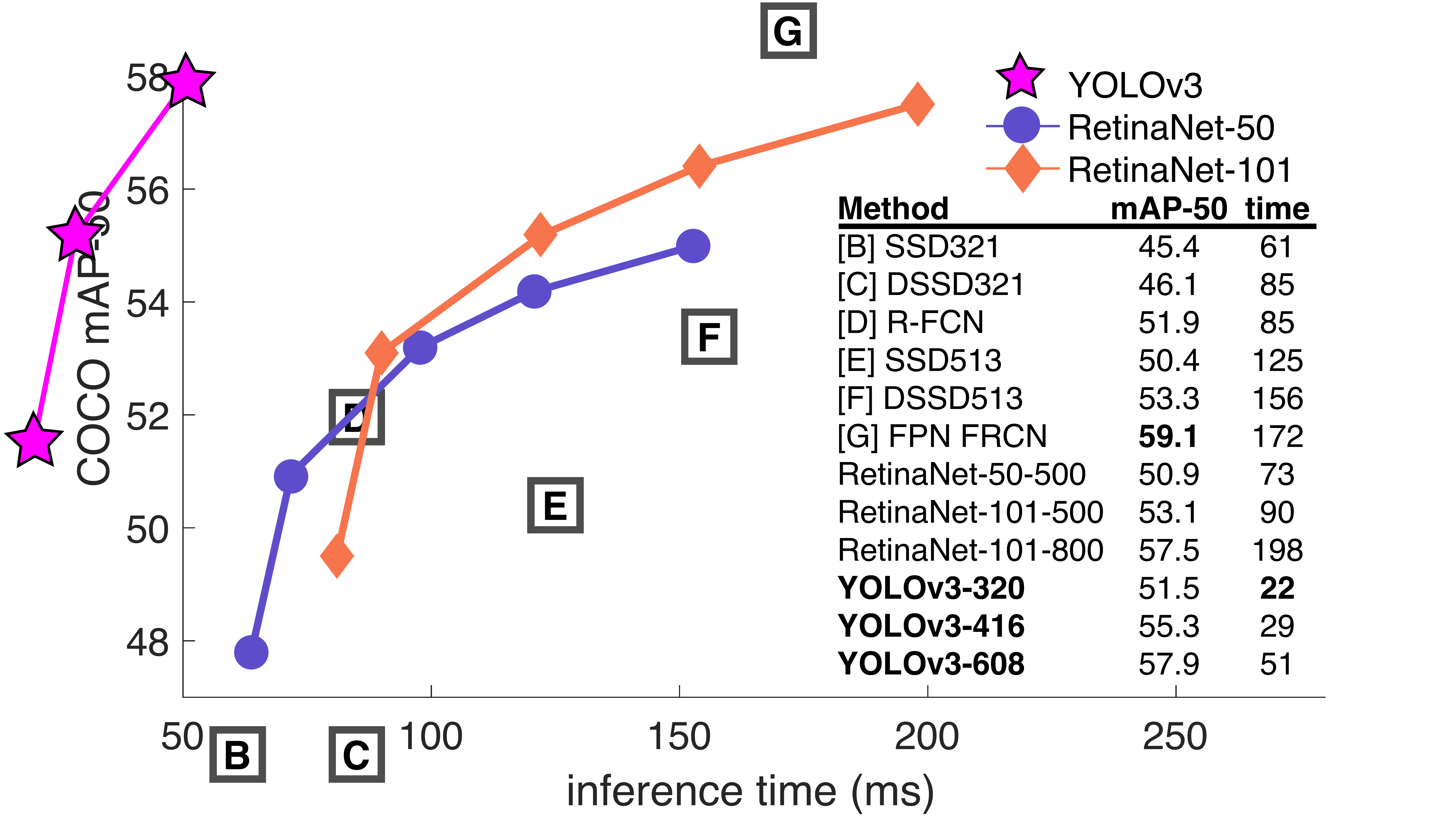}
\caption{Again adapted from the \cite{focal}, this time displaying speed/accuracy tradeoff on the mAP at .5 IOU metric. You can tell YOLOv3 is good because it's very high and far to the left. Can you cite your own paper? Guess who's going to try, this guy $\rightarrow$ \cite{yolov3}. Oh, I forgot, we also fix a data loading bug in YOLOv2, that helped by like 2 mAP. Just sneaking this in here to not throw off layout.}
\label{chart}
\end{figure*}

\section{Things We Tried That Didn't Work}

We tried lots of stuff while we were working on YOLOv3. A lot of it didn't work. Here's the stuff we can remember.

\textbf{Anchor box $x,y$ offset predictions.} We tried using the normal anchor box prediction mechanism where you predict the $x,y$ offset as a multiple of the box width or height using a linear activation. We found this formulation decreased model stability and didn't work very well.

\textbf{Linear $x,y$ predictions instead of logistic.} We tried using a linear activation to directly predict the $x,y$ offset instead of the logistic activation. This led to a couple point drop in mAP.

\textbf{Focal loss.} We tried using focal loss. It dropped our mAP about 2 points. YOLOv3 may already be robust to the problem focal loss is trying to solve because it has separate objectness predictions and conditional class predictions. Thus for most examples there is no loss from the class predictions? Or something? We aren't totally sure.

\textbf{Dual IOU thresholds and truth assignment.} Faster R-CNN uses two IOU thresholds during training. If a prediction overlaps the ground truth by .7 it is as a positive example, by $[.3 - .7]$ it is ignored, less than .3 for all ground truth objects it is a negative example. We tried a similar strategy but couldn't get good results.

We quite like our current formulation, it seems to be at a local optima at least. It is possible that some of these techniques could eventually produce good results, perhaps they just need some tuning to stabilize the training.

\section{What This All Means}

YOLOv3 is a good detector. It's fast, it's accurate. It's not as great on the COCO average AP between .5 and .95 IOU metric. But it's very good on the old detection metric of .5 IOU.

Why did we switch metrics anyway? The original COCO paper just has this cryptic sentence: ``A full discussion of evaluation metrics will be added once the evaluation server is complete''. Russakovsky et al report that  that humans have a hard time distinguishing an IOU of .3 from .5! ``Training humans to visually inspect a bounding box with IOU of 0.3 and distinguish it from one with IOU 0.5 is surprisingly difficult.'' \cite{russakovsky2015best} If humans have a hard time telling the difference, how much does it matter?

But maybe a better question is: ``What are we going to do with these detectors now that we have them?'' A lot of the people doing this research are at Google and Facebook. I guess at least we know the technology is in good hands and definitely won't be used to harvest your personal information and sell it to.... wait, you're saying that's exactly what it will be used for?? Oh.

Well the other people heavily funding vision research are the military and they've never done anything horrible like killing lots of people with new technology oh wait.....\footnote{The author is funded by the Office of Naval Research and Google.}

I have a lot of hope that most of the people using computer vision are just doing happy, good stuff with it, like counting the number of zebras in a national park \cite{parham2017animal}, or tracking their cat as it wanders around their house \cite{scott_2017}. But computer vision is already being put to questionable use and as researchers we have a responsibility to at least consider the harm our work might be doing and think of ways to mitigate it. We owe the world that much.

In closing, do not @ me. (Because I finally quit Twitter).

\newpage

{\small
\bibliographystyle{ieee}
\bibliography{references}

\begin{thebibliography}{10}\itemsep=-1pt

\bibitem{analogy}
Analogy.
\newblock {\em Wikipedia}, Mar 2018.

\bibitem{pascal}
M.~Everingham, L.~Van~Gool, C.~K. Williams, J.~Winn, and A.~Zisserman.
\newblock The pascal visual object classes (voc) challenge.
\newblock {\em International journal of computer vision}, 88(2):303--338, 2010.

\bibitem{fu2017dssd}
C.-Y. Fu, W.~Liu, A.~Ranga, A.~Tyagi, and A.~C. Berg.
\newblock Dssd: Deconvolutional single shot detector.
\newblock {\em arXiv preprint arXiv:1701.06659}, 2017.

\bibitem{gordon2017iqa}
D.~Gordon, A.~Kembhavi, M.~Rastegari, J.~Redmon, D.~Fox, and A.~Farhadi.
\newblock Iqa: Visual question answering in interactive environments.
\newblock {\em arXiv preprint arXiv:1712.03316}, 2017.

\bibitem{resnet}
K.~He, X.~Zhang, S.~Ren, and J.~Sun.
\newblock Deep residual learning for image recognition.
\newblock In {\em Proceedings of the IEEE conference on computer vision and
  pattern recognition}, pages 770--778, 2016.

\bibitem{huang2017speed}
J.~Huang, V.~Rathod, C.~Sun, M.~Zhu, A.~Korattikara, A.~Fathi, I.~Fischer,
  Z.~Wojna, Y.~Song, S.~Guadarrama, et~al.
\newblock Speed/accuracy trade-offs for modern convolutional object detectors.

\bibitem{openimages}
I.~Krasin, T.~Duerig, N.~Alldrin, V.~Ferrari, S.~Abu-El-Haija, A.~Kuznetsova,
  H.~Rom, J.~Uijlings, S.~Popov, A.~Veit, S.~Belongie, V.~Gomes, A.~Gupta,
  C.~Sun, G.~Chechik, D.~Cai, Z.~Feng, D.~Narayanan, and K.~Murphy.
\newblock Openimages: A public dataset for large-scale multi-label and
  multi-class image classification.
\newblock {\em Dataset available from https://github.com/openimages}, 2017.

\bibitem{lin2017feature}
T.-Y. Lin, P.~Dollar, R.~Girshick, K.~He, B.~Hariharan, and S.~Belongie.
\newblock Feature pyramid networks for object detection.
\newblock In {\em Proceedings of the IEEE Conference on Computer Vision and
  Pattern Recognition}, pages 2117--2125, 2017.

\bibitem{focal}
T.-Y. Lin, P.~Goyal, R.~Girshick, K.~He, and P.~Doll{\'a}r.
\newblock Focal loss for dense object detection.
\newblock {\em arXiv preprint arXiv:1708.02002}, 2017.

\bibitem{lin2014microsoft}
T.-Y. Lin, M.~Maire, S.~Belongie, J.~Hays, P.~Perona, D.~Ramanan,
  P.~Doll{\'a}r, and C.~L. Zitnick.
\newblock Microsoft coco: Common objects in context.
\newblock In {\em European conference on computer vision}, pages 740--755.
  Springer, 2014.

\bibitem{liu2016ssd}
W.~Liu, D.~Anguelov, D.~Erhan, C.~Szegedy, S.~Reed, C.-Y. Fu, and A.~C. Berg.
\newblock Ssd: Single shot multibox detector.
\newblock In {\em European conference on computer vision}, pages 21--37.
  Springer, 2016.

\bibitem{newton1687}
I.~Newton.
\newblock {\em Philosophiae naturalis principia mathematica}.
\newblock William Dawson \& Sons Ltd., London, 1687.

\bibitem{parham2017animal}
J.~Parham, J.~Crall, C.~Stewart, T.~Berger-Wolf, and D.~Rubenstein.
\newblock Animal population censusing at scale with citizen science and
  photographic identification.
\newblock 2017.

\bibitem{darknet13}
J.~Redmon.
\newblock Darknet: Open source neural networks in c.
\newblock \url{http://pjreddie.com/darknet/}, 2013--2016.

\bibitem{redmon2017yolo9000}
J.~Redmon and A.~Farhadi.
\newblock Yolo9000: Better, faster, stronger.
\newblock In {\em Computer Vision and Pattern Recognition (CVPR), 2017 IEEE
  Conference on}, pages 6517--6525. IEEE, 2017.

\bibitem{yolov3}
J.~Redmon and A.~Farhadi.
\newblock Yolov3: An incremental improvement.
\newblock {\em arXiv}, 2018.

\bibitem{ren2015faster}
S.~Ren, K.~He, R.~Girshick, and J.~Sun.
\newblock Faster r-cnn: Towards real-time object detection with region proposal
  networks.
\newblock {\em arXiv preprint arXiv:1506.01497}, 2015.

\bibitem{russakovsky2015best}
O.~Russakovsky, L.-J. Li, and L.~Fei-Fei.
\newblock Best of both worlds: human-machine collaboration for object
  annotation.
\newblock In {\em Proceedings of the IEEE Conference on Computer Vision and
  Pattern Recognition}, pages 2121--2131, 2015.

\bibitem{scott_2017}
M.~Scott.
\newblock Smart camera gimbal bot – scanlime:027, Dec 2017.

\bibitem{shrivastava2016beyond}
A.~Shrivastava, R.~Sukthankar, J.~Malik, and A.~Gupta.
\newblock Beyond skip connections: Top-down modulation for object detection.
\newblock {\em arXiv preprint arXiv:1612.06851}, 2016.

\bibitem{szegedy2017inception}
C.~Szegedy, S.~Ioffe, V.~Vanhoucke, and A.~A. Alemi.
\newblock Inception-v4, inception-resnet and the impact of residual connections
  on learning.
\newblock 2017.

\end{thebibliography}
}

\begin{figure*}[]
\vspace{-4mm}
\includegraphics[width=\linewidth]{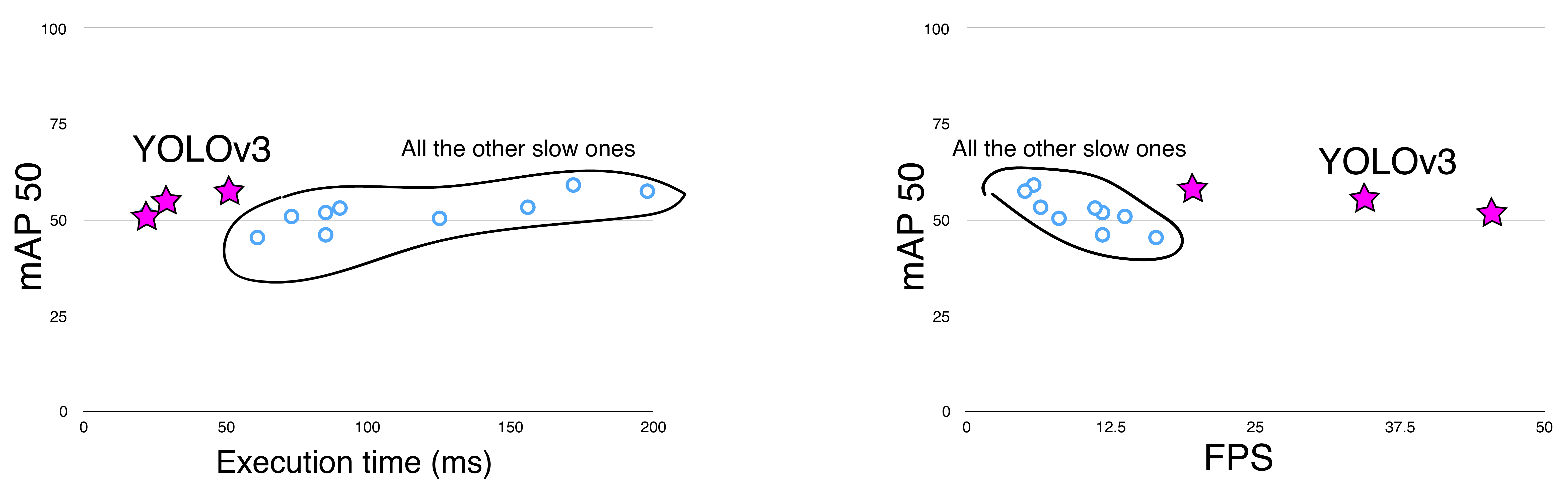}
\caption{Zero-axis charts are probably more intellectually honest... and we can still screw with the variables to make ourselves look good!}
\label{chart}
\end{figure*}

\newpage

\section*{Rebuttal}
\small

We would like to thank the Reddit commenters, labmates, emailers, and passing shouts in the hallway for their lovely, heartfelt words. If you, like me, are reviewing for ICCV then we know you probably have 37 other papers you could be reading that you'll invariably put off until the last week and then have some legend in the field email you about how you really should finish those reviews execept it won't entirely be clear what they're saying and maybe they're from the future? Anyway, this paper won't have become what it will in time be without all the work your past selves will have done also in the past but only a little bit further forward, not like all the way until now forward. And if you tweeted about it I wouldn't know. Just sayin.

Reviewer \#2 AKA Dan Grossman (lol blinding who does that) insists that I point out here that our graphs have not one but two non-zero origins. You're absolutely right Dan, that's because it looks way better than admitting to ourselves that we're all just here battling over 2-3\% mAP. But here are the requested graphs. I threw in one with FPS too because we look just like super good when we plot on FPS.


Reviewer \#4 AKA JudasAdventus on Reddit writes ``Entertaining read but the arguments against the MSCOCO metrics seem a bit weak''. Well, I always knew you would be the one to turn on me Judas. You know how when you work on a project and it only comes out alright so you have to figure out some way to justify how what you did actually was pretty cool? I was basically trying to do that and I lashed out at the COCO metrics a little bit. But now that I've staked out this hill I may as well die on it.

See here's the thing, mAP is already sort of broken so an update to it should maybe address some of the issues with it or at least justify why the updated version is better in some way. And that's the big thing I took issue with was the lack of justification. For \textsc{Pascal} VOC, the IOU threshold was ''set deliberately low to account for inaccuracies in bounding boxes in the ground truth data`` \cite{pascal}. Does COCO have better labelling than VOC? This is definitely possible since COCO has segmentation masks maybe the labels are more trustworthy and thus we aren't as worried about inaccuracy. But again, my problem was the lack of justification.


The COCO metric emphasizes better bounding boxes but that emphasis must mean it de-emphasizes something else, in this case classification accuracy. Is there a good reason to think that more precise bounding boxes are more important than better classification? A miss-classified example is much more obvious than a bounding box that is slightly shifted.

mAP is already screwed up because all that matters is per-class rank ordering. For example, if your test set only has these two images then according to mAP two detectors that produce these results are JUST AS GOOD:

\begin{figure}[h]
\includegraphics[width=\linewidth]{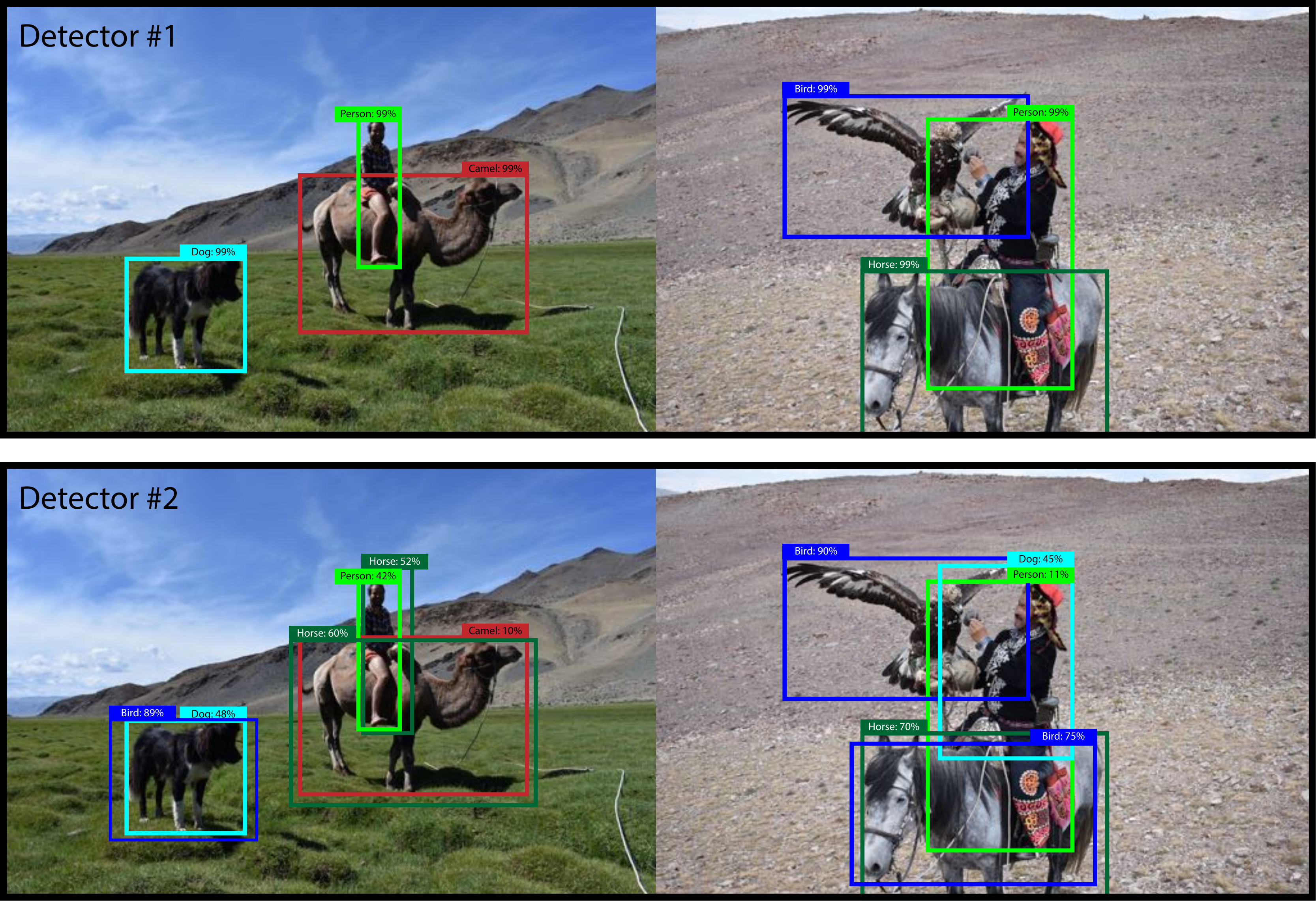}
\caption{These two hypothetical detectors are perfect according to mAP over these two images. They are both perfect. Totally equal.}
\label{chart}
\end{figure}

Now this is OBVIOUSLY an over-exaggeration of the problems with mAP but I guess my newly retconned point is that there are such obvious discrepancies between what people in the ``real world'' would care about and our current metrics that I think if we're going to come up with new metrics we should focus on these discrepancies. Also, like, it's already mean average precision, what do we even call the COCO metric, average mean average precision?

Here's a proposal, what people actually care about is given an image and a detector, how well will the detector find and classify objects in the image. What about getting rid of the per-class AP and just doing a global average precision? Or doing an AP calculation per-image and averaging over that?

Boxes are stupid anyway though, I'm probably a true believer in masks except I can't get YOLO to learn them.

\end{document}


\title{Supplementary Material for IQA: Visual Question Answering in Interactive Environments}
\maketitle

\section{Using Real Object Detection}
\begin{table}[!htbp]
\begin{center}
\resizebox{\columnwidth}{!}{
\begin{tabular}{|l|l|l|l|}
 \hline
 \multicolumn{4}{|c|}{Accuracy of QA} \\
 \hline
       Model & Existence &Counting & Spatial Relationship\\
 \hline
Random & 50 & 25 & 50 \\
A3C with no object detections & 56.9 & 26.42 & 59.1 \\
\textbf{A3C with YOLO object detection} & \textbf{54.29} & \textbf{26.78} & \textbf{55.36} \\
A3C with oracle object detections & 59.5 & 27.1 & 66.2 \\
\textbf{HIMN with YOLO object detection} & \textbf{63.39} & \textbf{35.89} & \textbf{57.14} \\
HIMN with oracle object detection & 69.8 & 32.2 & 65.6 \\
HIMN with oracle object detection and navigator & 73.03 & 45.35 & 71.42 \\
 \hline
\end{tabular}
}
\end{center}
\caption{This tables compares the test accuracy of question answering across different models.}
\label{table:supp_accuracy}
\end{table}

\begin{table}[!htbp]
\begin{center}
\resizebox{\textwidth}{!}{
\begin{tabular}{ |l|l|l|l|l|l|l|l|l|  }
 \hline
 \multicolumn{9}{|c|}{Accuracy of QA Per Answer} \\
 \hline
       Model & Existence & Existence & Counting & Counting & Counting & Counting & Spatial & Spatial \\
       & (N) & (Y) & (0) & (1) & (2) & (3) & Relation (N) & Relation (Y) \\
 \hline
Always Answer Most Likely Value & 57 & 0 & 27 & 0 & 0 & 0 & 52 & 0 \\
A3C with no object detections & 99.69 & 0.83 & 20.81 & 33.04 & 26.53 & 26.32 & 55.02 & 64.94 \\
A3C with YOLO object detection & 84.91 & 14.05 & 41.22 & 35.4 & 42.07 & 44.81 & 39.72 & 42.77 \\
A3C with oracle object detections & 100 & 1.92 & 34.29 & 11.11 & 35.14 & 23.08 & 64.21 & 68.09 \\
HIMN & 92.41 & 40 & 35.14 & 33.93 & 32.17 & 27.52 & 64.44 & 66.67 \\
HIMN with YOLO object detection & 92.14 & 25.62 & 48.32 & 53.57 & 48.30 & 50.66 & 50.93 & 52.89 \\
HIMN with oracle navigator & 100 & 37.6 & 36.24 & 53.57 & 51.02 & 42.11 & 70.82 & 72.29  \\
 \hline
\end{tabular}
}
\end{center}
\caption{Results for each question category broken down by each possible answer.}
\label{table:supp_breakdown}
\end{table}

The modular nature of \modelshort\ allows us to easily swap out and swap in controllers with different architectures. We replace ground truth object sensing (provided by the AI2-THOR framework) with an object detection algorithm (YOLO V2).  We fine-tune YOLO V2 \cite{redmon2016yolo9000} on the AI2-THOR training scenes to show it examples of classes that are not in the MSCOCO dataset such as bread and cabinet. We estimate the depth of an object using the FRCN depth estimation network \cite{laina2016deeper} and project the probabilities of the detected objects onto the ground plane. These detection probabilities are incorporated into the spatial memory using a moving average update rule.

To fairly compare against A3C, we also train the A3C model with YOLO object detection outputs rather than ground truth detections. Table~\ref{table:supp_accuracy} (comparable to Table 2 in the paper) shows the performance of \modelshort\ with YOLO compared to A3C with YOLO. \modelshort\ significantly outperforms this baseline and also outperforms A3C with ground truth object detections on 2 of the 3 question types. \modelshort\ is presumably able to learn robustness to detection noise because it can directly encode the YOLO object probability outputs into the spatial map and incorporate past observations from the same locations much more directly.

We further separate the accuracies for different answers to explore potential biases in the questions as well as in the model behaviors, shown in Table~\ref{table:supp_breakdown}. We find that the questions are mostly balanced, yet Existence is slightly more likely to be false than true. This is a result of filtering out Existence questions where the object is placed in an unobservable location. Because of this, we notice that the A3C models, rather than learn how to explore the environment, simply exploits the bias in the questions. Our model, on the other hand is still quite likely to get true existence questions correct.

\section{Network Architecture}
The full HIMN network can be broken into several networks for navigation, planning, and answering. Their architectures are as follows: \\

\noindent \textbf{Navigation Network} \\
Inputs: 
\begin{itemize}[itemsep=0pt]
    \item Current image at $ 300 \times 300 \times 3 $ resolution
    \item Previous Action One-Hot Vector
    \item Destination
\end{itemize}
Layers:
\begin{itemize}[itemsep=0pt]
    \item Conv: $64 \times 7 \times 7$ kernels, stride 2, ELU activation
    \item Max Pool: $2 \times 2$, stride 2
    \item Conv: $128 \times 5 \times 5$ kernels, stride 1, ELU activation
    \item Max Pool: $2 \times 2$, stride 2
    \item Conv: $256 \times 3 \times 3$ kernels, stride 1, ELU activation
    \item Conv: $256 \times 3 \times 3$ kernels, stride 1, ELU activation
    \item Conv: $256 \times 3 \times 3$ kernels, stride 1, ELU activation
    \item Max Pool: $2 \times 2$, stride 2
    \item Conv: $512 \times 3 \times 3$ kernels, stride 1, ELU activation
    \item Conv: $512 \times 3 \times 3$ kernels, stride 1, ELU activation
    \item Conv: $512 \times 3 \times 3$ kernels, stride 1, ELU activation
    \item Max Pool: $2 \times 2$, stride 2
    \item FC1: Fully Connected on conv output: 1024 units, ELU activation
    \item FC2: Fully Connected on action one-hot: 32 units, ELU activation
    \item FC-Concat: Concatenate (FC1, FC2)
    \item GRU: 1024 units
    \item GRU-Concat: Concatenate(FC-Concat, GRU)
    \item Spatial GRU: $32 \times 5 \times 5$
    \item \textbf{Output} Path Weight: Conv: $1 \times 1 \times 1$, stride 1, activation = $min(max(1, 5 * e^{x}), 200)$
    \item Crop: Spatial GRU at Destination with $5 \times 5$ padding
    \item Conv: $32 \times 3 \times 3$ kernels, stride 1, ELU activation
    \item Conv: $32 \times 3 \times 3$ kernels, stride 1, ELU activation
    \item \textbf{Output} Terminal: Fully Connected, no activation.
\end{itemize}

\noindent \textbf{Planner Network} \\
Inputs: 
\begin{itemize}[itemsep=0pt]
    \item Current image at $ 300 \times 300 \times 3 $ resolution
    \item Semantic map at $RoomW \times RoomH \times NumClasses + 5$
    \item Previous Action One-Hot Vector
    \item Question Encoding
\end{itemize}
Layers:
\begin{itemize}[itemsep=0pt]
    \item Conv: $32 \times 7 \times 7$ kernels, stride 2, ReLU activation
    \item Max Pool: $2 \times 2$, stride 2
    \item Conv: $64 \times 5 \times 5$ kernels, stride 1, ReLU activation
    \item Max Pool: $2 \times 2$, stride 2
    \item Conv: $128 \times 3 \times 3$ kernels, stride 1, ReLU activation
    \item Max Pool: $2 \times 2$, stride 2
    \item Conv: $256 \times 3 \times 3$ kernels, stride 1, ReLU activation
    \item FC1: Fully Connected on conv output: 1024 units, ELU activation
    \item FC2: Fully Connected on action one-hot: 32 units, ELU activation
    \item FC-Concat: Concatenate (FC1, FC2)
    \item GRU: 1024 units
    \item GRU-Concat: Concatenate(FC-Concat, GRU)
    \\
    \item FCQ: Fully connected on Question Encoding: 64 units, ELU activation
    \item Tile FCQ: $RoomW \times RoomH$
    \item Concatenate(Semantic Map, Tile FCQ)
    \item Conv: $64 \times 1 \times 1$ kernels, stride 1, ELU activation
    \item Conv: $64 \times 11 \times 11$ kernels, stride 1, ELU activation
    \item Conv: $128 \times 3 \times 3$ kernels, stride 1, ELU activation
    \item Conv: $256 \times 3 \times 3$ kernels, stride 1, ELU activation
    \item Semantic Features: Conv: $256 \times 3 \times 3$ kernels, stride 1, ELU activation
    \item \textbf{Output}: Viability: Conv: $1 \times 1 \times 1$, stride 1, no activation
    \item Crop: $1 \times 1$ at Semantic Features(current spatial location)
    \item Concatenate (GRU-Concat, Crop)
    \item \textbf{Output}: $V_{action}$: Fully Connected on Concatenate
    \item \textbf{Output}: $\pi_{action}$ Fully connected on Concatenate: 6 units
\end{itemize}

\noindent \textbf{Answerer Network} \\
Inputs: 
\begin{itemize}[itemsep=0pt]
    \item Semantic map at $RoomW \times RoomH \times NumClasses + 5$
    \item Question Encoding
\end{itemize}
Layers:
\begin{itemize}[itemsep=0pt]
    \item Tile Question: $RoomW \times RoomH$
    \item Concatenate: (Semantic Map, Tile Question)
    \item Conv: $64 \times 1 \times 1$ kernels, stride 1, ELU activation
    \item Conv: $128 \times 3 \times 3$ kernels, stride 2, ELU activation
    \item Max Pool: $2 \times 2$, stride 2
    \item Conv: $128 \times 3 \times 3$ kernels, stride 2, ELU activation
    \item Max Pool: $2 \times 2$, stride 2
    \item Conv: $128 \times 3 \times 3$ kernels, stride 2, ELU activation
    \item Max Pool: $2 \times 2$, stride 2
    \item Spatial Sum
    \item Fully Connected: 128 units, ELU activation
    \item Fully Connected: 128 units, ELU activation
    \item \textbf{Output}: Answer: Fully Connected: NumAnswerClasses, no activation
    \item \textbf{Output}: $V_{answer}$: Fully Connected, no activation
    \item \textbf{Output}: $\pi_{answer}$: Fully Connected, no activation
\end{itemize}

\section{Training details}
To train the navigation network, we used the ADAM optimization algorithm with the learning rate of $10^{-4}$, and default Tensorflow constants. We trained with a batch size of 256 for 200,000 iterations. To train the planner and answerer, we use the RMSProp optimization algorithm with a learning rate of $10^{-3}$. We use the learning curriculum described in the paper for 5 million iterations of A3C (5 million distinct interactions with the environment).

{\small
\bibliographystyle{ieee}
\bibliography{00_references}
}